\documentclass[conference]{IEEEtran}
\IEEEoverridecommandlockouts
\usepackage{cite}
\usepackage{amsmath,amssymb,amsfonts}
\usepackage{algorithmic}
\usepackage{graphicx}
\usepackage{textcomp}
\usepackage{xcolor}
\def\BibTeX{{\rm B\kern-.05em{\sc i\kern-.025em b}\kern-.08em
    T\kern-.1667em\lower.7ex\hbox{E}\kern-.125emX}}
\begin{document}

\makeatletter
\newcommand{\linebreakand}{%
  \end{@IEEEauthorhalign}
  \hfill\mbox{}\par
  \mbox{}\hfill\begin{@IEEEauthorhalign}
}
\makeatother

\title{Development of a Cacao Disease Identification and Management App Using Deep Learning}

\author{\IEEEauthorblockN{Zaldy O. Pagaduan}
\IEEEauthorblockA{\textit{School of Engineering and Architecture} \\
\textit{Ateneo de Davao University}\\
Davao City, Philippines \\
zopagaduanjr@addu.edu.ph}
\and
\IEEEauthorblockN{Jason T. Occidental*}
\IEEEauthorblockA{\textit{School of Engineering and Architecture} \\
\textit{Ateneo de Davao University}\\
Davao City, Philippines \\
jtoccidental@addu.edu.ph}
\linebreakand
\IEEEauthorblockN{Nathaniel N. Duro}
\IEEEauthorblockA{\textit{School of Engineering and Architecture} \\
\textit{Ateneo de Davao University}\\
Davao City, Philippines \\
nnduro@addu.edu.ph}
\and
\IEEEauthorblockN{Dexielito B. Badilles}
\IEEEauthorblockA{\textit{School of Engineering and Architecture} \\
\textit{Ateneo de Davao University}\\
Davao City, Philippines \\
dbbadilles@addu.edu.ph}
\and
\IEEEauthorblockN{Eleonor V. Palconit}
\IEEEauthorblockA{\textit{School of Engineering and Architecture} \\
\textit{Ateneo de Davao University}\\
Davao City, Philippines \\
evpalconit@addu.edu.ph}
}

\maketitle

\begin{abstract}
Smallholder cacao producers often rely on outdated farming techniques and face significant challenges from pests and diseases, unlike larger plantations with more resources and expertise. In the Philippines, cacao farmers have limited access to data, information, and good agricultural practices. This study addresses these issues by developing a mobile application for cacao disease identification and management that functions offline, enabling use in remote areas where farms are mostly located. The core of the system is a deep learning model trained to identify cacao diseases accurately. The trained model is integrated into the mobile app to support farmers in field diagnosis. The disease identification model achieved a validation accuracy of 96.93\%\, while the model for detecting cacao black pod infection levels achieved 79.49\%\ validation accuracy. Field testing of the application showed an agreement rate of 84.2\%\ compared with expert cacao technician assessments. This approach empowers smallholder farmers by providing accessible, technology-enabled tools to improve cacao crop health and productivity.
\end{abstract}

\begin{IEEEkeywords}
Cacao Dataset, Cacao Black Pod Rot, Cacao Pod Borer, Disease Identification, Deep Learning, AI in Agriculture
\end{IEEEkeywords}

\section{Introduction}
\subsection{Background}
The global cacao shortage is projected to fall short by over 1 million metric tons in 2025, creating a supply gap in the market. This is due to several factors, such as weather, diseases, and pests affecting the major producing countries \cite{b1}. 75\%\ of global cacao beans come from Africa \cite{b2}, which is battling drought, pests, and diseases \cite{b3}. High occurrence of pests and diseases, old age of cacao trees, and lack of soil nutrients are the primary causes of low cacao bean production in Africa \cite{b4}. Indonesia, another major player in the global cacao market, is moving from planting cacao to oil palm, which is a more profitable crop. The cause of Indonesia's shift is the negligence of the government in providing sufficient agronomic assistance on cacao \cite{b5}.
The infestation of pests and diseases in traditional cacao-producing countries led to a shortage in worldwide cacao production. With the global market shortage of cacao, everyone is searching for different cacao providers. Thus, the Philippines has a big opportunity to be part of the international cacao industry \cite{b6}.
The Philippine cacao industry is currently facing a disease management threat, with losses of about 30\%\ of cacao pods and 10\%\ tree mortality annually due to diseases \cite{b7}. In 2016, government and private agencies convened to better understand, detect, and control cacao diseases. According to Dr. Alvindia \cite{b8} of the Philippine Center for Postharvest Development and Mechanization (PHILMECH), early detection is crucial for the control of the disease.
By the end of 2018, 80 million cacao seedlings had been distributed and planted across the Philippines. Given the data, at least 40 to 80 thousand metric tons of cacao beans were expected to be produced, but the total cacao beans produced in 2018 were only 7 thousand metric tons \cite{b9}. Given the data, the seedlings, after several years, failed to produce the minimum cacao beans they can produce because the country's production is still low. According to Turtur \cite{b10}, growers lack technical assistance in improving the survival rates of planted seedlings.
A possible modern solution for the early detection of diseases is through automatic detection. It is necessary to integrate these automated services into a smartphone so that farmers can apply this technology in remote places \cite{b12}.

\subsection{Problem Statement}
Cacao farmers in the Philippines have limited access to relevant data, information, and knowledge of good agricultural practices. Good farming practices include integrated pest management, which could contribute to the cacao plant's susceptibility to pests and diseases if ignored \cite{b11}. The Philippines' four major cacao insect pests and diseases include cacao pod borer, cacao mirid bug, black pod rot, and vascular dieback \cite{b7,b13}. To better manage the pests and diseases on their farms, farmers need to be able to recognize the symptoms, understand the causes, and know how the pests and disease organisms operate \cite{b14}. 
Finally, the cacao industry is dominated by smallholder producers using outdated farming techniques. Smallholder producers combating pests and diseases face more obstacles compared to large plantations, which have funds and professional expertise to tackle these problems \cite{b7,b11,b15}. Several researchers have used Support Vector Machine in identifying cacao diseases \cite{b16,b17,b18}. However, in the case of cacao diseases, there still lacks a deep learning approach, which has already proven reliable results in the identification of other crop species and diseases \cite{b19,b20,b21,b22,b23}.

\subsection{Significance of the Study}
This study seeks to improve cacao production globally, considering that early detection and management of cacao diseases would help constitute a high-quality, sustainable cocoa production. In the Philippines, this would help fight the low production of cocoa due to diseases, eventually introducing the Philippines as an essential player in the cocoa-chocolate Global Value Chain. This study will further advertise the Philippines’ Department of Trade and Industry (DTI) and the Department of Agriculture's (DA’s) 2017-2022 Philippine Cacao Industry Roadmap's goal of producing 100,000 MT of dried fermented beans. This study can also support new farmers interested in planting cacao but who fear pests and diseases will deplete their investment in cacao farming. Lastly, it could improve the production of cocoa to meet the growing demand locally and internationally.
This study demonstrates the application of deep learning in image classification. Its higher accuracy and breakthroughs in image recognition are beneficial for disease identification, especially in the agricultural sector. Finally, it will showcase the capabilities of modern cellphones to handle complex model processing.

\subsection{Objectives of the Study}
The purpose of this study was to develop a disease identification and management application using deep learning. Specifically, this research aimed to:
\begin{itemize}
\item Gather the required number of datasets for each prevalent cacao disease in Davao City, Philippines.
\item Train a model that classifies cacao diseases and identifies the level of infection.
\item Develop a mobile application that can detect and classify cacao disease locally.
\item Create an automated management solution that can treat the identified disease.
\end{itemize}

\section{Materials and Methods}
\subsection{Research design and locale}
The study implemented a developmental research design focused on designing, developing, and evaluating a system that can identify cacao diseases and provide management solutions. Existing studies focused on cacao disease detection are limited due to a lack of data. They also infer that the current methods of detecting plant diseases are relatively difficult to operate, time-consuming, and require expert technicians. Thus, the research seeks to improve the current state of cacao disease detection by collecting the data needed to create an accurate model. The model was integrated into a mobile phone locally to provide quick results and suggest disease management solutions collected from cacao experts.
Data collection areas were done at three locations in Davao City, Philippines, particularly in the KVT cacao nursery and farm in Catalunan Grande of Talomo District, a private farm in Barangay Talomo River of Calinan District, and a private farm in Barangay Cadalian of Baguio District. KVT farm has four hectares of land filled with fruit-bearing cacao trees. The private farm in Barangay Talomo River has around 4000 trees, and the private farm in Barangay Cadalian has 700 cacao trees. 

\subsection{System Flow and Design}
Fig.~\ref{fig1} shows that the first process in the training phase involves the collection of the datasets. Around 1000 images were collected for each cacao disease and infection level. After the collection of datasets, data augmentation techniques are used to increase the dataset. Images were taken with varying angles and photo settings to minimize bias for machine learning. The process of detecting a specific disease also follows a particular manner; if the condition is manifesting in the fruit pod, all images of the disease will focus on the fruit pod.
\begin{figure}[htbp]
\centerline{\includegraphics[width=0.5\textwidth]{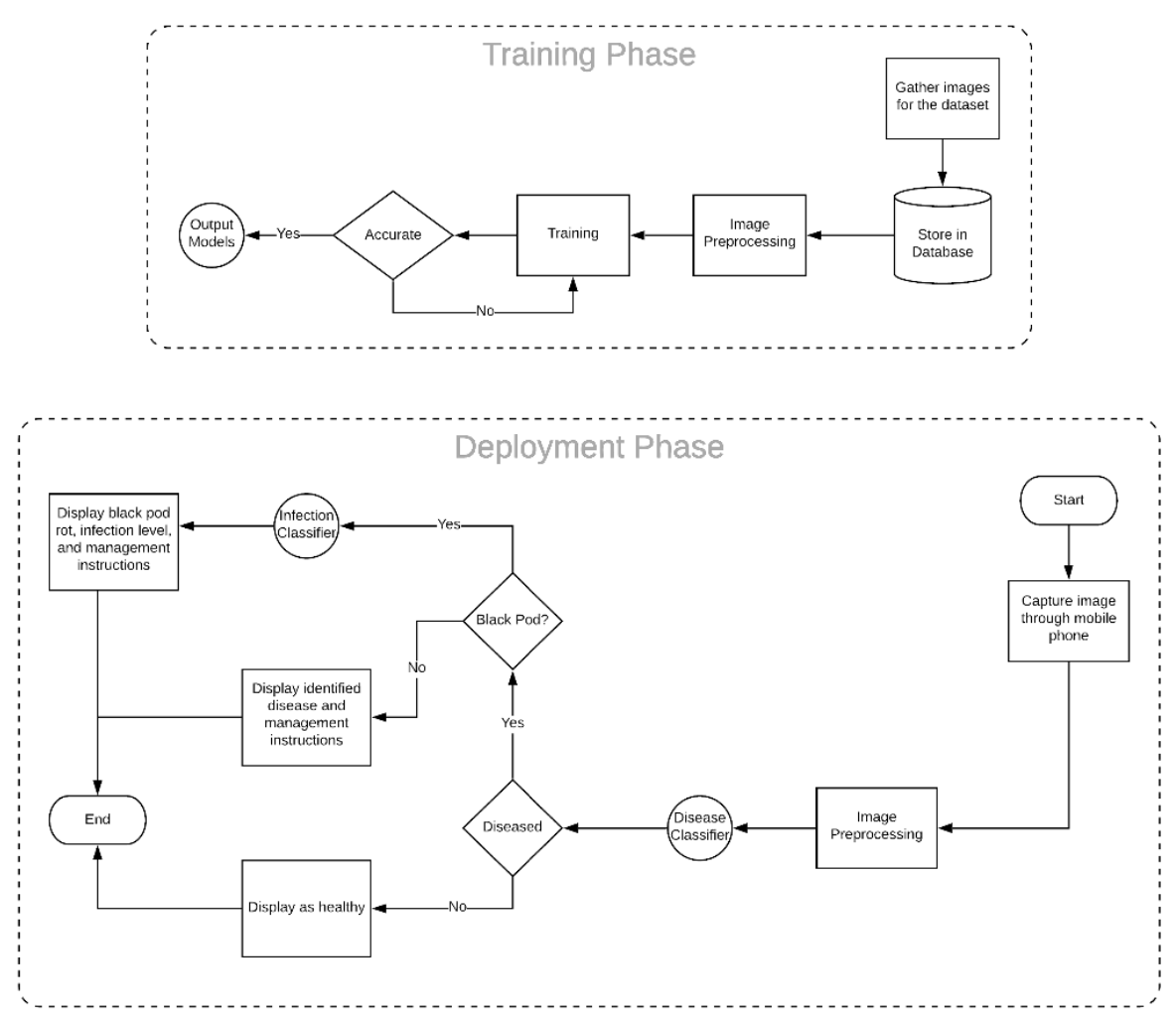}}
\caption{Cacao disease identification and management system flow.}
\label{fig1}
\end{figure}

After training thousands of images and epochs using deep learning architectures, accuracy and validation accuracy were measured. If the accuracy achieved does not meet the satisfactory level, the training phase repeats with adjusted parameters. The trained model with satisfactory metrics was saved.
After training, the model was exported into a format suitable for storing extensive collections of multidimensional numeric arrays. The exported model was converted into a FlatBuffer file, a dynamic, cross-platform serialization library. This file is necessary for mobile devices to use the trained model. The converted model will then be deployed to the next phase.
The deployment phase involved the development of the mobile application. The application is responsible for getting permission to handle the camera and image gallery, as well as capturing and choosing the image to be identified by the model. The model will process the image locally and display its specific disease, if it has any. If the particular disease is a black pod, the image will undergo another model that checks the black pod's infection level. The application then displays the recommended management practices corresponding to the disease detected.

\subsubsection{System design}
Fig.~\ref{fig2} shows that the Cacao disease dataset is used to train a classification model using a workstation with a discrete GPU. The trained model is integrated into a mobile application that allows users to take pictures of cacao for classification and shows disease management recommendations. The development of the mobile app required a specific version of an operating system and a camera. For the operating system, Android devices required Android 8.1, while Apple devices required iOS 12.0 or above.
\begin{figure}[htbp]
\centerline{\includegraphics[width=0.5\textwidth]{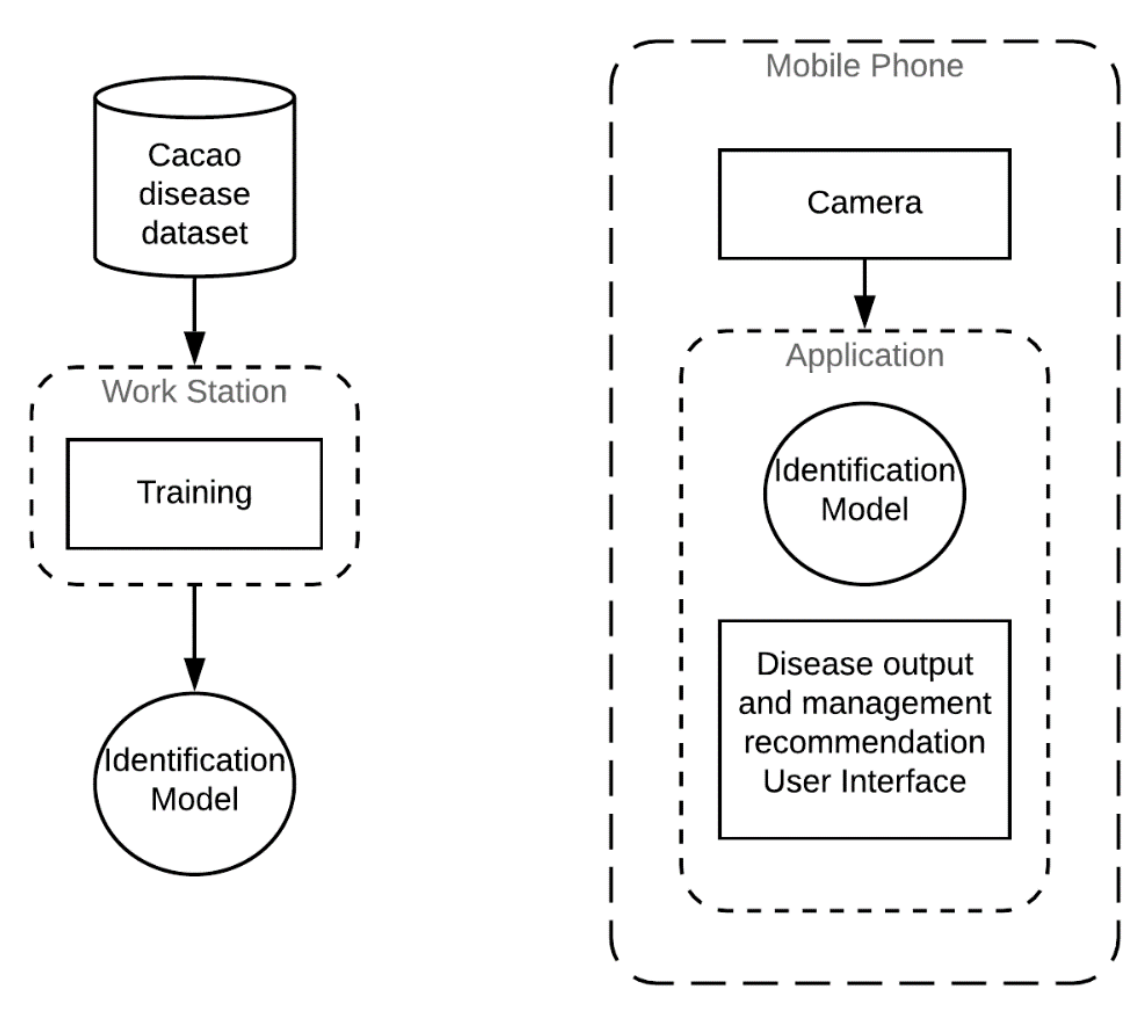}}
\caption{System overview.}
\label{fig2}
\end{figure}

\subsubsection{Deep learning model}
Fig.~\ref{fig3} presents an overview of designing a deep learning model. After the required dataset was gathered, the image was resized and divided across two sets, named the training set and the test set. The training set was much larger compared to the testing set, as it trained the model to learn the features of a diseased cacao. The training set has also undergone data augmentation to extend the dataset, as neural networks perform better with more data involved.
\begin{figure}[htbp]
\centerline{\includegraphics[width=0.5\textwidth]{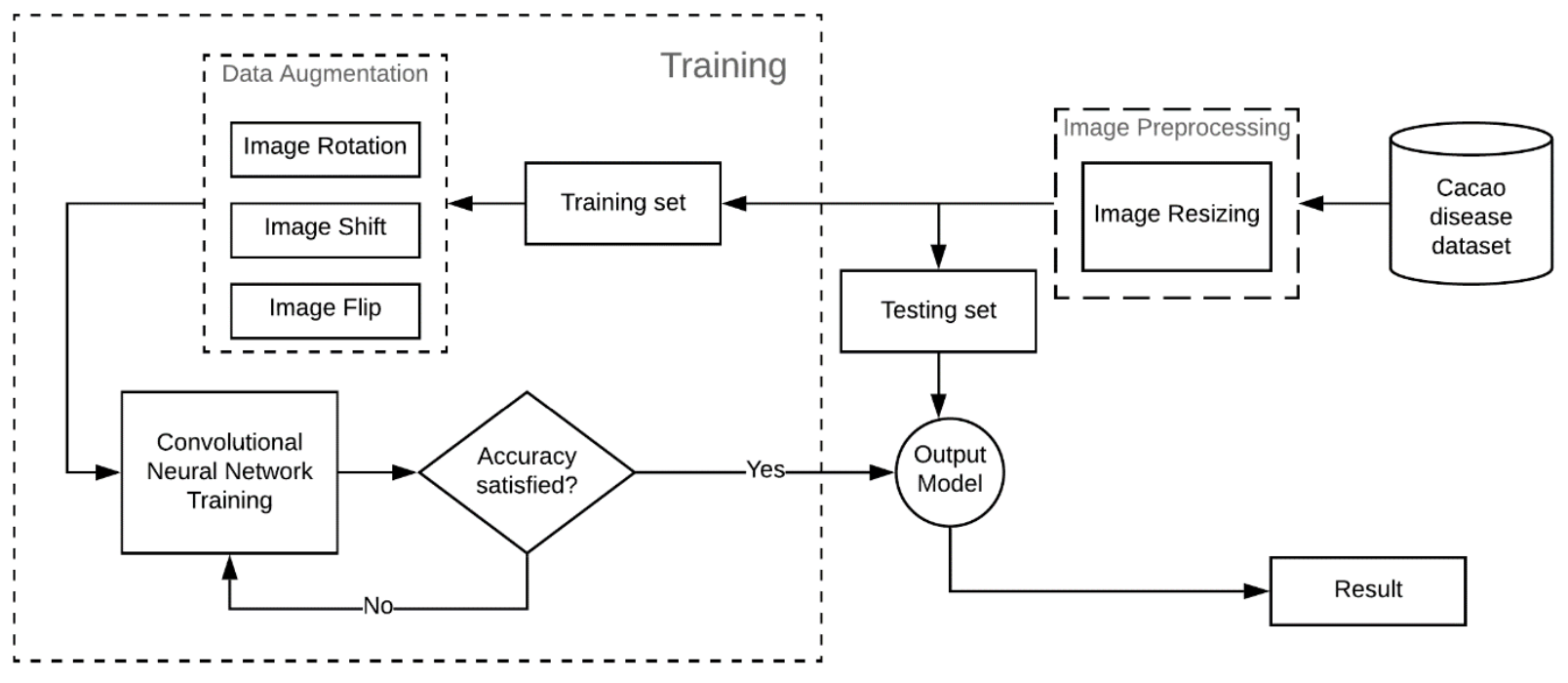}}
\caption{Model overview.}
\label{fig3}
\end{figure}

\subsubsection{Data cleaning}
The creation of automated tools and processes was vital in the cleaning process. The cleaning process removed images with blurred images or photos of foreign objects other than cacao. After removing the noticeable faulty images, the remaining data passes through a Python script that automatically resizes and renames the picture.
It proceeds to the initial labeling of images based on the description of the cacao technician. This labeling procedure also removed pods that do not fall under any category. Labeling process is shown on Fig.~\ref{figa}Lastly, the list of images was organized by category into their respective directory folders using a Python script. The Python libraries used for data cleaning are os, Python Imaging Library, and shutil.

\begin{figure}[htbp]
\centerline{\includegraphics[width=0.5\textwidth]{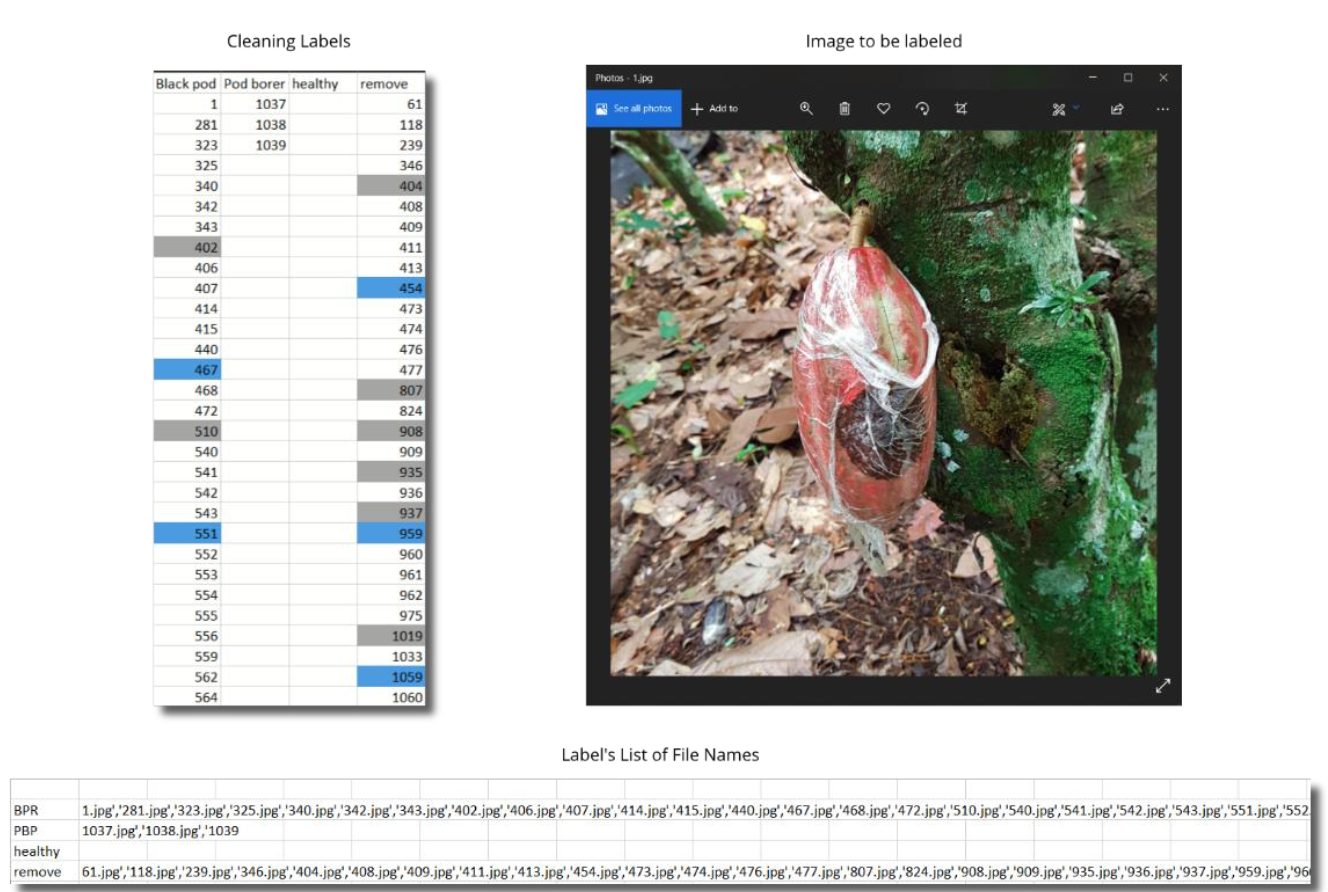}}
\caption{Breakdown of the data cleaning process.}
\label{figa}
\end{figure}

\subsubsection{Model building}
The proposed convolutional neural network architecture was based on EfficientNet, a novel model scaling method that scales up the layers efficiently. This scaling technique helps avoid model saturation when only one hyperparameter is increased \cite{b24}. Fig.~\ref{figb} illustrates the compound scaling method of EfficientNet.

\begin{figure}[htbp]
\centerline{\includegraphics[width=0.5\textwidth]{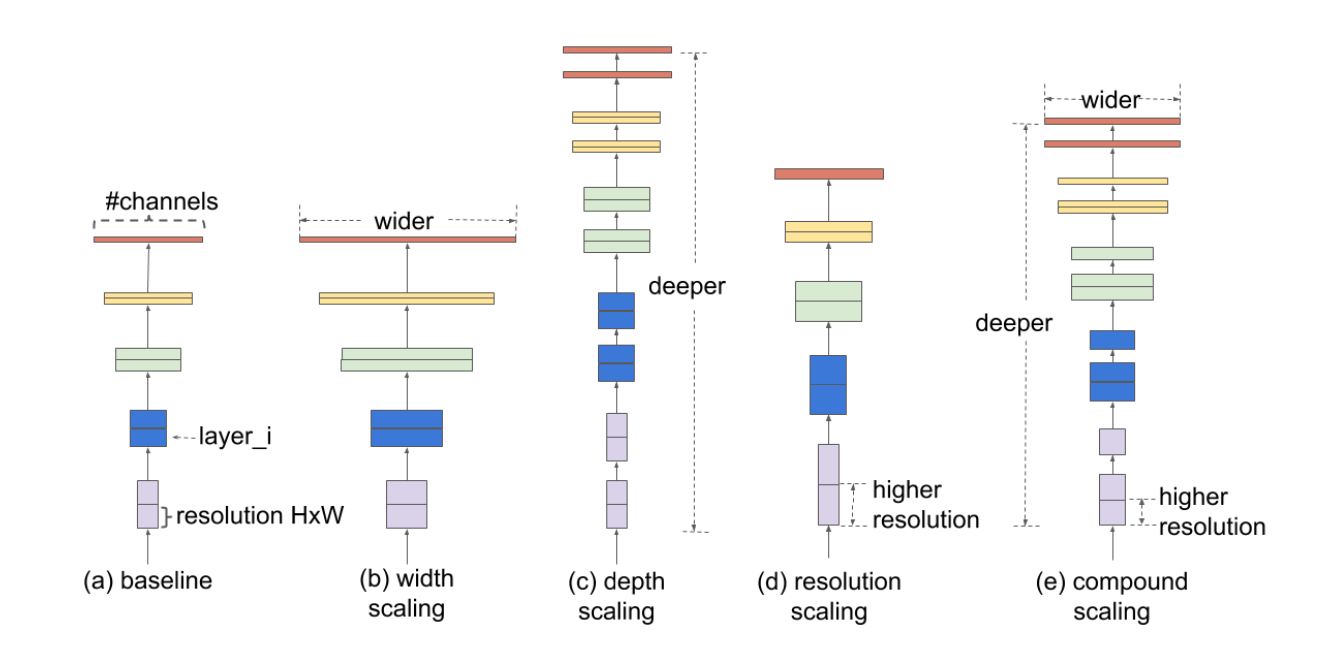}}
\caption{EfficientNet compound scaling.}
\label{figb}
\end{figure}

\subsubsection{Mobile application design}
The software application was developed using Flutter in Android Studio using the Dart programming language. 
This service handles the camera, image gallery, capturing, and choosing the image to be identified by the model. The model can process the image without an internet connection. It starts by loading the cacao disease model and classifying whether the image of a cacao pod is black pod rot, healthy, or pod borer. If the image is classified as black pod rot, the black pod rot level classification model is loaded to identify the infection level of the black pod rot. The result is stored as a HiveObject using HiveDB, an encrypted and fast key-value NoSQL database.
The application also provides disease management recommendations. This service depends on the disease determined by the model. The recommended management solutions are gathered from multiple cacao experts and plant pathologists. The management recommendation is divided into  three (3) parts: the treatment, the symptoms, and the sources. The management recommendation is stored as a JSON file.
The Rectified Linear Unit (ReLU) was used to activate the convolutional layer. ReLU increases the model's nonlinearity without altering the convolutional layer's receptive field by simply replacing all the negative values with 0. The formula is represented as:

\begin{equation}
F(x) = max (0, x)\label{eq}
\end{equation}

Average pooling is also used to downsize the layer to reduce spatial dimensions without affecting the depth. Reducing the feature map reduces training time and computation cost and controls overfitting. The proposed model is a multi-class classifier, which requires a softmax function to be used as the final layer. Mathematically, the softmax function can be represented as a function that maps:

\begin{equation}
S(a): \mathbb{R}^N \to \mathbb{R}^N\label{eq}
\end{equation}

The research utilized cloud services such as Amazon SageMaker, Google Colab, and Kaggle. During the project implementation, the model was trained on SageMaker, Colab, and Kaggle for the last phase of the design project. These cloud services helped ease the demand for a workstation since model building requires extensive GPU and CPU computational power.

\section{Results and Discussion}
\subsection{Data Collection}
The dataset collection happened over nine days or eight weekends. The collection focused on capturing both healthy and diseased cacao pod images. A total of 4,980 images were collected over the nine days of data collection. The breakdown of each set of images are shown on Table ~\ref{tab1} based on the location where they were captured.

\begin{table}[htbp]
\caption{Data collection points and dates}
\begin{center}
\begin{tabular}{|c|c|c|}
\hline
\textbf{Date} & \textbf{Place} & \textbf{No.of Raw Photos} \\
\hline
2 July 2020 & KVY Farm Nursery & 335  \\
12 Sept. 2020 & Novela Farms & 917  \\
19 Sept. 2020 & Novela Farms & 725  \\
20 Sept. 2020 & Novela Farms & 662  \\
10 Oct. 2020 & Novela Farms & 1,212  \\
18 Oct. 2020 & Novela Farms & 503  \\
10 Jan. 2021 & Private Farm & 246  \\
16 Jan. 2021 & Private Farm & 224  \\
23 Jan. 2021 & Private Farm & 156  \\
\hline
\multicolumn{2}{|c|}{\textbf{Total}} & 4,980  \\
\hline
\end{tabular}
\label{tab1}
\end{center}
\end{table}

\subsection{Cleaning and Validation}
After cleaning and sorting the images collected, the dataset is now ready to be reviewed by a cacao expert. Consultation is necessary to minimize bias made during the collection and cleaning of data. The cacao expert also provided insights and tips on how to make the dataset more robust. After validation, the number of images were 4390. 

\subsection{Model building}
The first phase of the model building was to augment parts of the dataset, as the resulting dataset was imbalanced. The next step is to assemble, define, and use compound scaling techniques for the model's layers, inspired by EfficientNet configuration. The last phase was to train the model to achieve acceptable accuracy, tuning multiple parameters during the process. After reaching a sufficient accuracy, the model is converted into a FlatBuffer file, a format readable by a smartphone.

\subsection{Training Accuracy}
For the cacao disease model at epoch 7, the training accuracy had reached 98.56\%\, while the validation accuracy, the accuracy of datasets never seen before, attained 96.93\%\. The model building automatically saves the best model and halts training when the model is no longer learning key features. Table~\ref{tab2} shows the precision, recall, f1-score, and support metrics for the best model.

\begin{table}[htbp]
\caption{Precision, Recall, F1-Score, and Support Metrics}
\begin{center}
\begin{tabular}{|c|c|c|c|c|}
\hline
& \textbf{Precision} & \textbf{Recall} & \textbf{F1-score} & \textbf{Support} \\
\hline
Black pod rot & 93.41\%\ & 89.95\%\ & 91.64\%\ & 189 \\
\hline
Pod Borer Pest & 65.62\%\ & 100\%\ & 79.25\%\ & 21  \\
\hline
Healthy & 97.14\%\ & 96.56\%\ & 96.85\%\ & 669   \\
\hline
\end{tabular}
\label{tab2}
\end{center}
\end{table}

The model-building process stopped at the 10th epoch, since during its last three epochs, the model has not learned anything new, as shown in Fig.~\ref{fig4}. The resulting model at epoch seven was then converted to a FlatBuffer file ready to be integrated with the mobile app.
\begin{figure}[htbp]
\centerline{\includegraphics[width=0.5\textwidth]{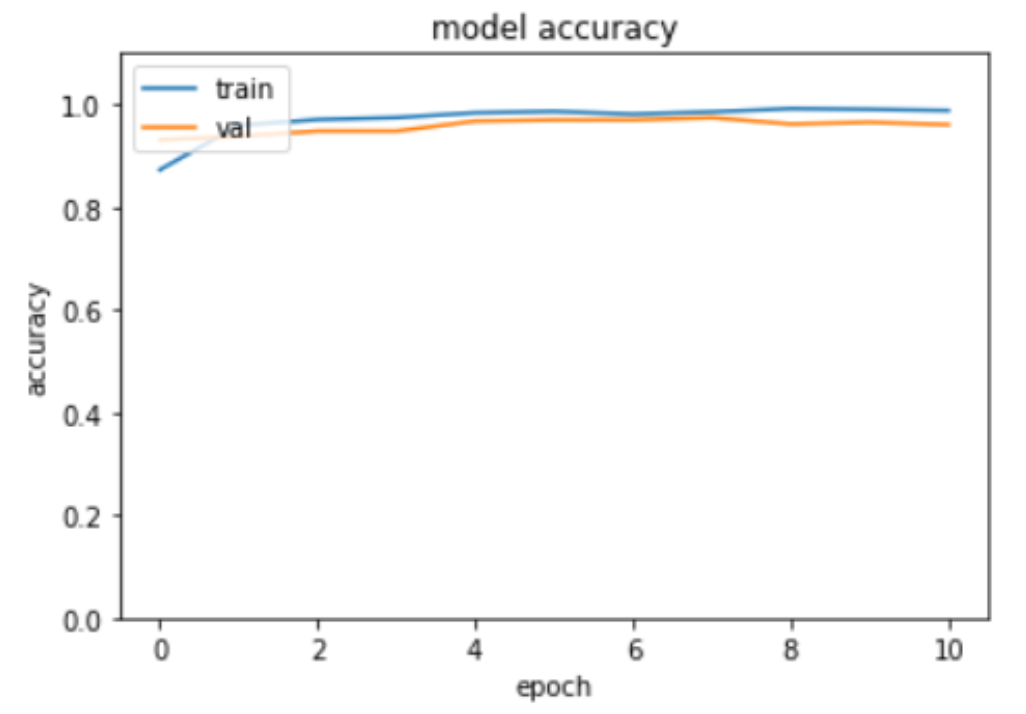}}
\caption{Accuracy flattens with increasing epoch.}
\label{fig4}
\end{figure}

The confusion matrix below displays the correct and incorrect predictions of the cacao disease model. The model performed relatively well, gaining 96.93\%\ despite the data being unbalanced.

\begin{figure}[htbp]
\centerline{\includegraphics[width=0.5\textwidth]{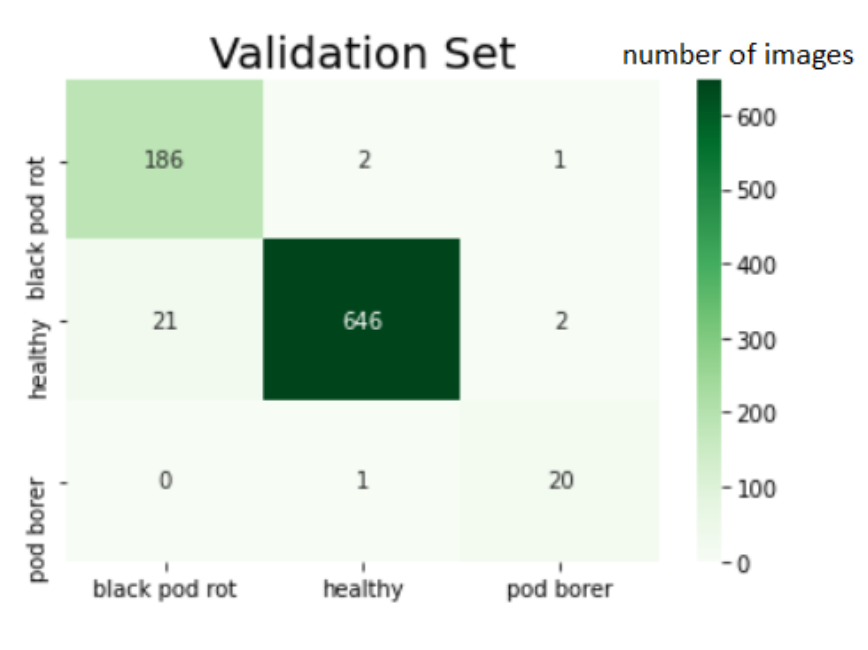}}
\caption{Black pod infection levels confusion matrix.}
\label{fig5}
\end{figure}

In comparison to other studies, the 2016 paper of Tan et al. \cite{b17} had 60\%\ accuracy as it correctly identified three out of five validation images. Additionally, the 2018 paper AuToDiDAC \cite{b16} has correctly identified four out of 10 validation images, resulting in a 40\%\ accuracy.

\subsection{Field testing}
For the real-time accuracy, 19 cacao pods were chosen in the field. The mobile app, together with the cacao technician, diagnosed and inferred on the cacao pods. The mobile app had 84.2\%\ overall similarity with the judgment of the cacao technician. The three (3) incorrect predictions were a healthy pod, a level 3 black pod rot, and a level 2 black pod rot, which the model classified as pod borer.
The mobile app had 4 out of 6 similar predictions with the cacao technician on black pod rot pods, while the AuToDiDAC had matched 4 out of 10 black pod rot \cite{b16}.

\begin{figure}[htbp]
\centerline{\includegraphics[width=0.5\textwidth]{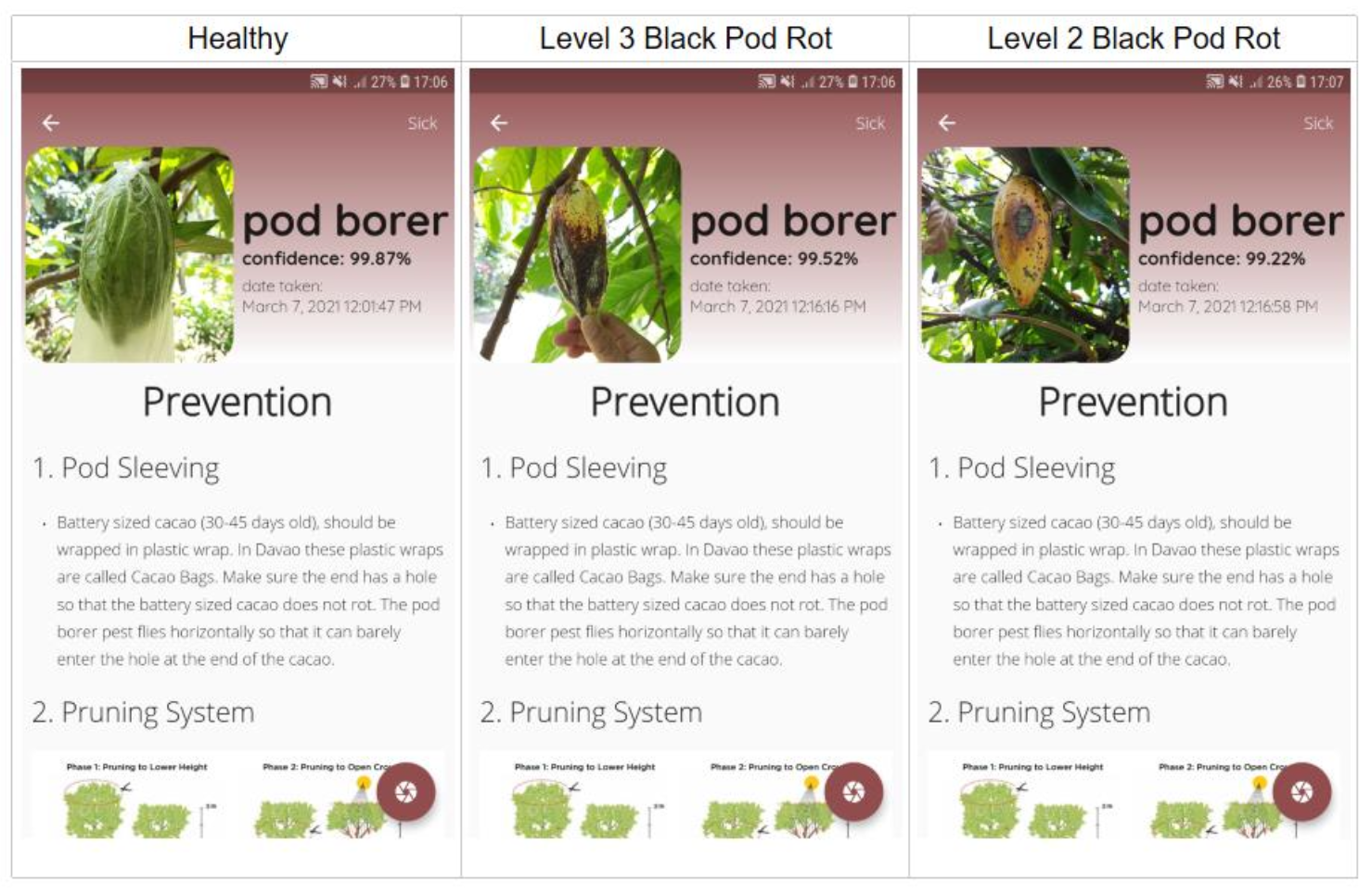}}
\caption{Incorrect prediction by the mobile application.}
\label{fig6}
\end{figure}

\section{Conclusion}
In this study, a deep learning approach to a real-world problem has been developed. The first step in this approach was the collection of multiple images of cacao pods. Next were the cleaning and verification of the collected datasets. Once verified, analysis and model building were done. The approach produced a deployable model ready to be utilized via a mobile application. 
In addition, the study has produced two datasets: cacao diseases and black pod levels, verified by a cacao technician. The study also constructed two models: one focuses on cacao diseases, and the other on black pod levels. 
Finally, the study produced a mobile application that can handle multiple models without relying on the cloud. Focusing on the locally deployed app is essential, as cacao farms in general have poor network coverage. Feedback from six farmers indicates that the app performs satisfactorily overall.

\section*{Acknowledgment}
The study acknowledges the support of the Offices of the Municipal Agricultural Officer in the Davao Region for their assistance in carrying out the project. Moreover, we thank the farms of Evangeline Novela, Val Turtur, and Hubert Sartorio for welcoming the researcher to their farms. Finally, to Mr. Dario Divino, who assisted in the validation of the datasets and the models developed for the study.

\end{document}